\documentclass{article}
\usepackage[legalpaper, portrait, margin=0.99in]{geometry}

\usepackage[english]{babel}

\usepackage{amsmath}
\usepackage{graphicx}
\usepackage[colorlinks=true, allcolors=blue]{hyperref}
\usepackage{authblk}
\usepackage{amsfonts} 
\usepackage{comment}

\title{Vectorized Adjoint Sensitivity Method for Graph Convolutional Neural Ordinary Differential Equations}
\author[1]{Jack Cai}
\affil[1]{University of Toronto, Division of Engineering Science}
\date{August 9, 2021}

\begin{document}
\maketitle

\begin{abstract}
This document, as the title stated, is meant to provide a vectorized implementation of adjoint dynamics calculation for Graph Convolutional Neural Ordinary Differential Equations (GCDE). The adjoint sensitivity method is the gradient approximation method for neural ODEs that replaces the back propagation. When implemented on libraries such as PyTorch or Tensorflow, the adjoint can be calculated by autograd functions without the need for a hand-derived formula. In applications such as edge computing and in memristor crossbars, however, autograds are not available, and therefore we need a vectorized derivation of adjoint dynamics to efficiently map the system on hardware. This document will go over the basics, then move on to derive the vectorized adjoint dynamics for GCDE.
\end{abstract}

\section{Introduction and Preliminaries}

Neural Ordinary Differential Equations (ODE) is a class of residual neural networks that frames the layer-wise propagation of hidden states into an initial value problem using differential equation solvers. 
\begin{equation}
h_{t+1} = h_{t} + f(h_{t}, \theta_{t})
\end{equation}
A typical residual network may have layer wise transformation of hidden states looks like Equation (1). If we add more layers and take smaller time step, then the entire problem can be framed as an ODE defined by Equation (2), and a solution at time $t_{1}$ defined by Equation (3). 
\begin{equation}
\frac{dh(t)}{dt} = f(h(t), t, \theta)
\end{equation}
\begin{equation}
h(t_{1}) = h(t_{0}) + \int_{t_{0}}^{t_{1}} f(h(t), t, \theta)dt
\end{equation}
Compared to conventional deep neural networks (DNNs), neural ODE uses less parameters applied more times to the hidden states, thereby achieves the depth of DNN while having higher memory efficiency. 

\subsection{Adjoint sensitivity method}
The adjoint sensitivity method is used for automatic differentiation for neural ODE. Unlike traditional backpropagation, a quantity named adjoint is calculated for each $t$ and another ODE solver is used to integrate the overall gradient. Let $L()$ be the scaler loss function for the neural ODE, the adjoint is defined as $a(t) = \frac{\partial L}{\partial h(t)}$, and its dynamic is given by Equation (4):
\begin{equation}
\frac{da(t)}{dt} = -a(t)^{T}\frac{\partial f(h(t), t, \theta)}{\partial h}
\end{equation}
The adjoint at each instant is calculated by a backward ODE solver, and the overall gradient of the parameters $\frac{\partial L}{\partial \theta}$ is given by Equation (5), which is integrated by another ODE solver:
\begin{equation}
\frac{\partial L}{\partial \theta} = -\int_{t_{1}}^{t_{0}} a(t)^{T}\frac{\partial f(h(t), t, \theta)}{\partial \theta}dt
\end{equation}
The proof and intuition behind adjoint sensitivity method is presented in the original paper. For the scope of this paper, we are only focused on how Equation (4) and (5) can be vectorized for GCDE. 

\subsection{GCDE}
GCDE, in short, is the neural ODE version of the state of the art graph learning method Graph Convolutional Neural Network (GCN). For GCN and GCDE, hidden state is represented by a matrix H of dimension $\mathbb{R}^{N\times C}$, representing $N$ nodes with $C$ features. Each layer of GCN or each step of GCDE involves node-wise exchange of information and convolution on nodes. The dynamic of GCDE (which is the same as the layer-wise propagation of GCN) is given by Equation (6), where $A\in\mathbb{R}^{N\times N}$ is symmetric that represents the graph topology and $W\in\mathbb{R}^{C\times C}$ represents the convolution filters:
\begin{equation}
\frac{dH(t)}{dt} = f(H(t), A, W) = ReLU(AH(t)W)
\end{equation}
To calculate the adjoint, we need to calculate calculate the Jacobian of the function $f(H(t), A, W)$, namely $\frac{\partial f(H(t), A, W)}{\partial H(t)}$ and $\frac{\partial f(H(t), A, W)}{\partial W}$. This poses us a challenge as we will have to unroll the matrix into vectors. While this can be done easily with built-in autograd functions in PyTorch and Tensorflow, it becomes messy when we write it out to obtain an analytical vectorized solution. 

\subsection{Matrix conventions}
Before we delve into deriving the adjoint dynamics, I would like to discuss some convention used in this document. All matrices are represented by capital letters, all vectors are represented by bold lower case letter, all entries within a matrix are represented by the $i$,$j$ subscripts which denotes the $i_{th}$ and $j_{th}$ entry of the matrix, and all entries within a vector are represented by a single subscript denoting the index. For example, $A$ is a matrix, $\mathbf{b}$ is a vector, $a_{1, 2}$ denotes the $1, 2$ entry of $A$, and $b_{2}$ denotes the second entry of $\mathbf{b}$. We use $``:"$ to denote the entire indices along a row or column, so that we represent the $i_{th}$ row of $A$ as $\mathbf{a}_{i,:}$, and we represent $j_{th}$ column of A as $\mathbf{a}_{:,j}$.

\subsection{Multivariable calculus conventions}
 
Let $f: \mathbb{R}^{n} \rightarrow \mathbb{R}^{m}$ where it takes in a vector $\mathbf{x} \in\mathbb{R}^{n}$ and output a vector $\mathbf{y} \in\mathbb{R}^{m}$, i.e.:

\begin{equation}
\begin{bmatrix}
    y_{1}       \\
    y_{2}      \\
    \vdots \\
    y_{m}      
\end{bmatrix}
= f(
\begin{bmatrix}
    x_{1} \\
    x_{2} \\
    \vdots \\
    x_{n} 
\end{bmatrix}
)
\end{equation}
Then the Jacobian matrix $J_{f} \in \mathbb{R}^{m\times n}$ is defined as:

\begin{equation}
J_{f} = f(
\begin{bmatrix}
    \frac{\partial y_{1}}{\partial x_{1}} & \frac{\partial y_{1}}{\partial x_{2}} &\hdots & \frac{\partial y_{1}}{\partial x_{n}} \\
    \frac{\partial y_{2}}{\partial x_{1}}  & \ddots &\hdots & \vdots \\
    \vdots & \vdots & \ddots & \vdots\\
    \frac{\partial y_{m}}{\partial x_{1}}  & \hdots & \hdots & \frac{\partial y_{m}}{\partial x_{n}} 
\end{bmatrix}
)
\end{equation}
Following this convention, let $f: \mathbb{R}^{n} \rightarrow \mathbb{R}^{m}$ and $g: \mathbb{R}^{m} \rightarrow \mathbb{R}^{p}$, then $g \circ f: \mathbb{R}^{n} \rightarrow \mathbb{R}^{p}$. The chain rule is defined as: 
\begin{equation}
    J_{g \circ f} = J_{g}\cdot J_{f}
\end{equation}

\subsection{Partial derivatives of a matrix}
Finding the Jacobian for GCDE is challenging not only because it is a composite function, but also because the input is a matrix, we need to unroll the matrix into vectors to match our matrix Jacobian convention. Let $roll()$ and $unroll()$ to be such operations:
\begin{equation}
    \mathbb{R}^{m\times n}\ni A = roll(\mathbf{a}), \mathbf{a}\in\mathbb{R}^{mn}
\end{equation}
\begin{equation}
    \mathbb{R}^{mn}\ni\mathbf{a} = unroll(A),  A\in\mathbb{R}^{m\times n}
\end{equation}
The example below illustrate how they can be used. Suppose $f: \mathbb{R}^{m\times n} \rightarrow \mathbb{R}^{p\times n}$ defined by $f(A) = BA$, where $B \in \mathbb{R}^{p\times m}$. Using Equation (10) and (11), we can convert $f$ into $g: \mathbb{R}^{mn}\rightarrow \mathbb{R}^{pn}$, such that:
\begin{equation}
    g(\mathbf{a}) = unroll(B(roll(\mathbf{a})))
\end{equation}
As a result, we can obtained a Jacobian matrix $J_{g} \in \mathbb{R}^{pn\times mn}$, which follows the convention defined in section 1.4. The way we unroll a matrix could be by rows or by columns -- it really depends on the situation.
\section{Adjoint derivation for GCDE}
The key of finding the Jacobian matrix for $f(H(t), A, W)$ is to find the Jacobian matrix for the three-matrix multiplication step, as the derivative for $ReLU$ is just an elementwise binary step function. So before we moving on solving the Jacobian matrix for the whole thing, let's consider the general case function $f(A) = XAY: \mathbb{R}^{n\times p}  \rightarrow \mathbb{R}^{m\times q}$ where $X\in\mathbb{R}^{m\times n}$, $A\in\mathbb{R}^{n\times p}$, and $Y\in\mathbb{R}^{p\times q}$. We can think of it as composite function $f(A) = h\circ g(A)$, where:
\begin{equation}
    g(A) = XA = B, B\in\mathbb{R}^{m\times p}
\end{equation}
\begin{equation}
h(B) = BY = C, C\in\mathbb{R}^{m\times q}
\end{equation}
According to our convention, we need to unroll the matrix into vectors in order to calculate the Jacobian matrix. Let $\widehat{g}$ and $\widehat{h}$ to denote them and $\mathbf{a} = unroll(A)$ and $\mathbf{b} = unroll(B)$:
\begin{equation}
\widehat{g}(\mathbf{a}) = unroll(X(roll(\mathbf{a})))
\end{equation}
\begin{equation}
\widehat{h}(\mathbf{b}) = unroll((roll(\mathbf{b}))Y)\end{equation}
\begin{center}
    $\widehat{f}(\mathbf{a}) = \widehat{h}\circ\widehat{g}(\mathbf{a})$
\end{center}
Hence, according to chain rule, the Jacobian matrix $J_{\widehat{h}\circ\widehat{g}} = J_{\widehat{h}}\cdot J_{\widehat{g}}$ and we will calculate it step by step below.

\subsection{Jacobian matrix $J_{\widehat{g}}$ for $g(A)=XA$}
We will start by breaking $X$ into rows and $A$ into columns. According to our convention outlined in section 1.3, Equation (13) becomes:

\begin{center}$g(A) = XA = \begin{bmatrix} \mathbf{x}_{1,:} \\ \vdots \\ \mathbf{x}_{1,:} \end{bmatrix} \begin{bmatrix} \mathbf{a}_{:,1} && \hdots && \mathbf{a}_{:,p} \end{bmatrix} $
\end{center}

\begin{center}$
= \begin{bmatrix}
    \mathbf{x}_{1,:} \cdot \mathbf{a}_{:,1} & \mathbf{x}_{1,:} \cdot \mathbf{a}_{:,2} &\hdots & \mathbf{x}_{1,:} \cdot \mathbf{a}_{:,p} \\
    \mathbf{x}_{2,:} \cdot \mathbf{a}_{:,1}  & \ddots &\hdots & \vdots \\
    \vdots & \vdots & \ddots & \vdots\\
    \mathbf{x}_{m,:} \cdot \mathbf{a}_{:,1}  & \hdots & \hdots & \mathbf{x}_{m,:} \cdot \mathbf{a}_{:,p}
\end{bmatrix}
= \begin{bmatrix}
    b_{1, 1} & b_{1,2} &\hdots & b_{1,p} \\
    b_{2,1}  & \ddots &\hdots & \vdots \\
    \vdots & \vdots & \ddots & \vdots\\
    b_{m,1}  & \hdots & \hdots & b_{m,p}
\end{bmatrix}
$
\end{center}
Since we used rows of $X$ and columns of $A$, for our convenience, we will unroll $B$ by rows and $A$ by columns. Therefore our Jacobian matrix $J_{\widehat{g}}$ becomes:
\begin{center}$
J_{\widehat{g}} = 
\begin{bmatrix}
    \frac{\partial b_{1,1}}{\partial a_{1,1}} & \frac{\partial b_{1,1}}{\partial a_{2,1}} &\hdots &  \frac{\partial b_{1,1}}{\partial a_{1,2}} &\hdots & \frac{\partial b_{1,1}}{\partial a_{n,p}} \\
    \frac{\partial b_{1, 2}}{\partial a_{1,1}}  & \ddots &\hdots & \frac{\partial b_{1,2}}{\partial a_{1,2}}& \hdots& \vdots \\
    \vdots & \vdots & \ddots & \vdots & \hdots & \vdots\\
    \frac{\partial b_{2,1}}{\partial a_{1,1}} & \vdots & \vdots & \ddots & \hdots & \vdots\\
    \vdots & \vdots & \vdots & \vdots & \ddots & \vdots\\
    \frac{\partial b_{m,p}}{\partial a_{1,1}}  & \hdots & \hdots & \hdots & \hdots & \frac{\partial b_{m,p}}{\partial a_{n,p}}
\end{bmatrix}
=
\begin{bmatrix}
    \frac{\partial \mathbf{x}_{1,:} \cdot \mathbf{a}_{:,1}}{\partial a_{1,1}} & \frac{\partial \mathbf{x}_{1,:} \cdot \mathbf{a}_{:,1}}{\partial a_{2,1}} &\hdots &  \frac{\partial \mathbf{x}_{1,:} \cdot \mathbf{a}_{:,1}}{\partial a_{1,2}} &\hdots & \frac{\partial \mathbf{x}_{1,:} \cdot \mathbf{a}_{:,1}}{\partial a_{n,p}} \\
    \frac{\partial \mathbf{x}_{1,:} \cdot \mathbf{a}_{:,2}}{\partial a_{1,1}}  & \ddots &\hdots & \frac{\partial \mathbf{x}_{1,:} \cdot \mathbf{a}_{:,2}}{\partial a_{1,2}}& \hdots& \vdots \\
    \vdots & \vdots & \ddots & \vdots & \hdots & \vdots\\
    \frac{\partial \mathbf{x}_{2,:} \cdot \mathbf{a}_{:,1}}{\partial a_{1,1}} & \vdots & \vdots & \ddots & \hdots & \vdots\\
    \vdots & \vdots & \vdots & \vdots & \ddots & \vdots\\
    \frac{\partial \mathbf{x}_{m,:} \cdot \mathbf{a}_{:,m}}{\partial a_{1,1}}  & \hdots & \hdots & \hdots & \hdots & \frac{\partial \mathbf{x}_{m,:} \cdot \mathbf{a}_{:,p}}{\partial a_{n,p}}
\end{bmatrix}
$
\end{center}
Notice that for $\frac{\partial \mathbf{x}_{1, :} \cdot \mathbf{a}_{:,1}}{\partial a_{2,1}}$, the partial derivative is non-zero because $\mathbf{x}_{1, :} \cdot \mathbf{a}_{:,1}$ depends on the entry $a_{2,1}$; whereas for $\frac{\partial \mathbf{x}_{1,:} \cdot \mathbf{a}_{:,2}}{\partial a_{1,1}}$, the partial derivative is zero because $a_{1,1}$ is not an entry inside the column vector $\mathbf{a}_{:, 2}$, therefore change in $a_{1,1}$ do not impact $\mathbf{x}_{1,:} \cdot \mathbf{a}_{:,2}$. \textit{The entry must be within the column vector to have a non-zero partial derivative}. Following this pattern, our Jacobian matrix $J_{\widehat{g}}$ becomes:
\begin{center}$
J_{\widehat{g}} = \begin{bmatrix}
    \frac{\partial \mathbf{x}_{1,:} \cdot \mathbf{a}_{:,1}}{\partial a_{1,1}} & \frac{\partial \mathbf{x}_{1,:} \cdot \mathbf{a}_{:,1}}{\partial a_{2,1}} &\hdots &  0 &\hdots & 0 \\
    0  & \ddots &\hdots & \frac{\partial \mathbf{x}_{1,:} \cdot \mathbf{a}_{:,2}}{\partial a_{1,2}}& \hdots& \vdots \\
    \vdots & \vdots & \ddots & \vdots & \hdots & \vdots\\
    \frac{\partial \mathbf{x}_{2,:} \cdot \mathbf{a}_{:,1}}{\partial a_{1,1}} & \vdots & \vdots & \ddots & \hdots & \vdots\\
    \vdots & \vdots & \vdots & \vdots & \ddots & \vdots\\
    0  & \hdots & \hdots & \hdots & \hdots & \frac{\partial \mathbf{x}_{m,:} \cdot \mathbf{a}_{:,p}}{\partial a_{n,p}}
\end{bmatrix}
$
\end{center}
Taking the derivative of the dot product, vectorize it and we get the Jacobian matrix $J_{\widehat{g}}$:
\begin{equation}
J_{\widehat{g}} = \begin{bmatrix}
    x_{1,1} &\hdots & x_{1,n}&  0 &\hdots & 0 \\
    0   &\ddots & 0 &x_{1,1}& \hdots& \vdots \\
    \vdots & \vdots & \ddots & \vdots & \hdots & \vdots\\
    x_{2,1}  & \vdots & x_{2,n} & \ddots & \hdots & \vdots\\
    \vdots & \vdots & \vdots & \vdots & \ddots & \vdots\\
    0  & \hdots & \hdots & \hdots & \hdots & x_{m,n}
\end{bmatrix} = \begin{bmatrix}
     \mathbf{x}_{1,:}  & &   &  \\
      & \mathbf{x}_{1,:} & &  \\
     &  & \ddots &\\
     &  &  & \mathbf{x}_{1,:}\\
    \mathbf{x}_{2,:}  & &   &  \\
      & \mathbf{x}_{2,:} & &  \\
     &  & \ddots &\\
     &  &  & \mathbf{x}_{2,:}\\
     & \hdots& \hdots &\\
     \mathbf{x}_{m,:}  & &   &  \\
      & \mathbf{x}_{m,:} & &  \\
     &  & \ddots &\\
     &  &  & \mathbf{x}_{m,:}
\end{bmatrix}
\in \mathbb{R}^{mp \times np}
\end{equation}

\subsection{Jacobian matrix $J_{\widehat{h}}$ for $h(B)=BY$}
Similarly to section 2.1, we will start by breaking $B$ and $Y$ into rows and columns. So Equation (14) becomes:

\begin{center}$
h(B) = BY = \begin{bmatrix} \mathbf{b}_{1,:} \\ \vdots \\ \mathbf{b}_{m,:} \end{bmatrix} 
\begin{bmatrix} \mathbf{y}_{:,1} && \hdots && \mathbf{y}_{:,q} \end{bmatrix} $
\end{center}

\begin{center}$
= \begin{bmatrix}
    \mathbf{b}_{1,:} \cdot \mathbf{y}_{:,1} & \mathbf{b}_{1,:} \cdot \mathbf{y}_{:,2} &\hdots & \mathbf{b}_{1,:} \cdot \mathbf{y}_{:,p} \\
    \mathbf{b}_{2,:} \cdot \mathbf{y}_{:,1}  & \ddots &\hdots & \vdots \\
    \vdots & \vdots & \ddots & \vdots\\
    \mathbf{b}_{m,:} \cdot \mathbf{y}_{:,1}  & \hdots & \hdots & \mathbf{b}_{m,:} \cdot \mathbf{y}_{:,p}
\end{bmatrix}
= \begin{bmatrix}
    c_{1,1} & c_{1,2} &\hdots & c_{1,p} \\
    c_{2,1}  & \ddots &\hdots & \vdots \\
    \vdots & \vdots & \ddots & \vdots\\
    c_{m,1}  & \hdots & \hdots & c_{m,p}
\end{bmatrix}
$
\end{center}
We will unroll B by rows and C by columns to construct our Jacobian matrix:
\begin{center}$
J_{\widehat{h}} = 
\begin{bmatrix}
    \frac{\partial c_{1,1}}{\partial b_{1,1}} & \frac{\partial c_{1,1}}{\partial b_{1,2}} &\hdots &  \frac{\partial c_{1,1}}{\partial b_{2,1}} &\hdots & \frac{\partial c_{1,1}}{\partial b_{m,p}} \\
    \frac{\partial c_{1,2}}{\partial b_{1,1}}  & \ddots &\hdots & \frac{\partial c_{1,2}}{\partial b_{2,1}}& \hdots& \vdots \\
    \vdots & \vdots & \ddots & \vdots & \hdots & \vdots\\
    \frac{\partial c_{2,1}}{\partial b_{1,1}} & \vdots & \vdots & \ddots & \hdots & \vdots\\
    \vdots & \vdots & \vdots & \vdots & \ddots & \vdots\\
    \frac{\partial c_{m,q}}{\partial b_{1,1}}  & \hdots & \hdots & \hdots & \hdots & \frac{\partial c_{m,q}}{\partial b_{m,p}}
\end{bmatrix}
=
\begin{bmatrix}
    \frac{\partial \mathbf{b}_{1,:} \cdot \mathbf{y}_{:,1} }{\partial b_{1,1}} & \frac{\partial \mathbf{b}_{1,:} \cdot \mathbf{y}_{:,1} }{\partial b_{1,2}} &\hdots &  \frac{\partial \mathbf{b}_{1,:} \cdot \mathbf{y}_{:,1} }{\partial b_{2,1}} &\hdots & \frac{\partial \mathbf{b}_{1,:} \cdot \mathbf{y}_{:,1} }{\partial b_{m,p}} \\
    \frac{\partial \mathbf{b}_{1,:} \cdot \mathbf{y}_{:,2} }{\partial b_{1,1}}  & \ddots &\hdots & \frac{\partial \mathbf{b}_{1,:} \cdot \mathbf{y}_{:,2} }{\partial b_{2,1}}& \hdots& \vdots \\
    \vdots & \vdots & \ddots & \vdots & \hdots & \vdots\\
    \frac{\partial \mathbf{b}_{2,:} \cdot \mathbf{y}_{:,1} }{\partial b_{1,1}} & \vdots & \vdots & \ddots & \hdots & \vdots\\
    \vdots & \vdots & \vdots & \vdots & \ddots & \vdots\\
    \frac{\partial \mathbf{b}_{m,:} \cdot \mathbf{y}_{:,q} }{\partial b_{1,1}}  & \hdots & \hdots & \hdots & \hdots & \frac{\partial \mathbf{b}_{m,:} \cdot \mathbf{y}_{:,q} }{\partial b_{m,p}}
\end{bmatrix}
$
\end{center}
By the pattern that non-pairing entry and row vectors have zero partial derivatives, the Jacobian matrix can be reduced to:
\begin{center}$
J_{\widehat{h}}
=
\begin{bmatrix}
    \frac{\partial \mathbf{b}_{1,:} \cdot \mathbf{y}_{:,1} }{\partial b_{1,1}} & \frac{\partial \mathbf{b}_{1,:} \cdot \mathbf{y}_{:,1} }{\partial b_{1,2}} &\hdots &  0 &\hdots & 0 \\
    \frac{\partial \mathbf{b}_{1,:} \cdot \mathbf{y}_{:,2} }{\partial b_{1,1}}  & \ddots &\hdots & 0 & \hdots& \vdots \\
    \vdots & \vdots & \ddots & \vdots & \hdots & \vdots\\
    0 & \vdots & \vdots & \ddots & \hdots & \vdots\\
    \vdots & \vdots & \vdots & \vdots & \ddots & \vdots\\
    0  & \hdots & \hdots & \hdots & \hdots & \frac{\partial \mathbf{b}_{m,:} \cdot \mathbf{y}_{:,q} }{\partial b_{m,p}}
\end{bmatrix}
$
\end{center}
Taking the partial derivatives of the dot product:
\begin{center}$
J_{\widehat{h}}
=
\begin{bmatrix}
    y_{1,1} & y_{2,1} &\hdots & y_{p,1}&  0 &\hdots & 0 \\
    y_{1,2}  & \vdots &\ddots  &y_{p,2}&0& \hdots& \vdots \\
    \vdots &\vdots& \vdots & \ddots & \hdots & \hdots & \vdots\\
    y_{1,q} & y_{2,q} & \vdots & y_{p,q}& 0 & \hdots & \vdots\\
    0& 0 & \vdots & 0& y_{1,1} & \hdots & \vdots\\
    0 & 0 & \vdots & 0& y_{1,2} & \hdots & \vdots\\
    \vdots&\vdots & \vdots & \vdots & \vdots & \ddots & \vdots\\
    0 & 0 & \vdots & 0& y_{1,q} & \hdots & \vdots\\
    \vdots&\vdots & \vdots & \vdots & \vdots & \ddots & \vdots\\
    0  &\hdots& \hdots & \hdots & \hdots & \hdots & y_{p,q}
\end{bmatrix}
$
\end{center}
Notice that the diagonal $q\times p$ blocks are $Y^T$, therefore:
\begin{equation}
J_{\widehat{h}}
=
\begin{bmatrix}
    Y^T \\ & Y^T \\ & & Y^T \\ & & & \ddots \\ & & & & Y^T
\end{bmatrix} \in \mathbb{R}^{mq \times mp}
\end{equation}

\subsection{Calculating $J_{\widehat{h}\circ\widehat{g}} $ from the chain rule}
By the chain rule, we calculate $J_{\widehat{h}\circ\widehat{g}} \in \mathbb{R}^{mq \times np}$ by $J_{\widehat{h}}\cdot J_{\widehat{g}}$:
\begin{center}$
J_{\widehat{h}}\cdot J_{\widehat{g}}
=
\begin{bmatrix}
    Y^T \\ & Y^T \\ & & Y^T \\ & & & \ddots \\ & & & & Y^T
\end{bmatrix} \cdot 
\begin{bmatrix}
     \mathbf{x}_{1,:}  & &   &  \\
      & \mathbf{x}_{1,:} & &  \\
     &  & \ddots &\\
     &  &  & \mathbf{x}_{1,:}\\
     & \hdots& \hdots &\\
     \mathbf{x}_{m,:}  & &   &  \\
      & \mathbf{x}_{m,:} & &  \\
     &  & \ddots &\\
     &  &  & \mathbf{x}_{m,:}
\end{bmatrix}
$
\end{center}
To simplify the matrix multiplication, we break done $Y^T$ to (note that $\mathbf{y}_{i,:}^T \in \mathbb{R}^{q\times1}$ is the transpose of the first row of $Y$):
\begin{center}$
Y^T=\begin{bmatrix}
    \mathbf{y}_{1,:}^T & \mathbf{y}_{2,:}^T & \hdots & \mathbf{y}_{p,:}^T
\end{bmatrix} 
$\end{center}
And we break the matrix multiplication blocks by blocks:
\begin{center}
    $J_{\widehat{h}}\cdot J_{\widehat{g}}=
    \begin{bmatrix}
    \begin{bmatrix}Y^T & \textbf{0}& \hdots& \textbf{0}\end{bmatrix} \cdot J_{\widehat{g}}\\
    \begin{bmatrix}\textbf{0}& Y^T& \hdots& \textbf{0}\end{bmatrix} \cdot J_{\widehat{g}}\\
    \hdots     \\
     \begin{bmatrix}\textbf{0} & \textbf{0}& \hdots& Y^T\end{bmatrix} \cdot J_{\widehat{g}}   
    \end{bmatrix} 
    $
\end{center}
Let's look at the first block:
\begin{equation}
\begin{bmatrix}Y^T & \textbf{0}& \hdots& \textbf{0}\end{bmatrix} \cdot J_{\widehat{g}} = 
    \begin{bmatrix}
    Y^T & \textbf{0}& \hdots& \textbf{0}
\end{bmatrix} \cdot 
\begin{bmatrix}
     \mathbf{x}_{1,:}  & &   &  \\
      & \mathbf{x}_{1,:} & &  \\
     &  & \ddots &\\
     &  &  & \mathbf{x}_{1,:}\\
     & \hdots& \hdots &\\
     \mathbf{x}_{m,:}  & &   &  \\
      & \mathbf{x}_{m,:} & &  \\
     &  & \ddots &\\
     &  &  & \mathbf{x}_{m,:}
\end{bmatrix}
\end{equation}
Note that since $Y^T$  $\in q\times p$, only the first $p$ rows of $J_{\widehat{g}}$ matters. Hence Equation (19) becomes: 
\begin{center}
    $Y^T \cdot
    \begin{bmatrix}
     \mathbf{x}_{1,:}  & &   &  \\
      & \mathbf{x}_{1,:} & &  \\
     &  & \ddots &\\
     &  &  & \mathbf{x}_{1,:}\\
\end{bmatrix}$
\end{center}

\begin{center}
    $=\begin{bmatrix}
    \mathbf{y}_{1,:}^T & \mathbf{y}_{2,:}^T & \hdots & \mathbf{y}_{p,:}^T
\end{bmatrix} \cdot
    \begin{bmatrix}
     \mathbf{x}_{1,:}  & &   &  \\
      & \mathbf{x}_{1,:} & &  \\
     &  & \ddots &\\
     &  &  & \mathbf{x}_{1,:}\\
\end{bmatrix}$
\end{center}
\begin{center}
    $= \begin{bmatrix}
     \mathbf{y}_{1,:}^T\cdot\mathbf{x}_{1,:}  &  \mathbf{y}_{2,:}^T\cdot\mathbf{x}_{1,:} & \hdots & \mathbf{y}_{p,:}^T\cdot\mathbf{x}_{1,:}\end{bmatrix} \in \mathbb{R}^{q\times pn}$
\end{center}
If we repeat for other blocks, we would get the Jacobian matrix for $J_{\widehat{h}\circ\widehat{g}} $ to be:

\begin{equation}
    J_{\widehat{h}\circ\widehat{g}} = \begin{bmatrix}
    \mathbf{y}_{1,:}^T\cdot\mathbf{x}_{1,:}  &  \mathbf{y}_{2,:}^T\cdot\mathbf{x}_{1,:} & \hdots & \mathbf{y}_{p,:}^T\cdot\mathbf{x}_{1,:} \\
    \mathbf{y}_{1,:}^T\cdot\mathbf{x}_{2,:}  &  \mathbf{y}_{2,:}^T\cdot\mathbf{x}_{2,:} & \hdots & \mathbf{y}_{p,:}^T\cdot\mathbf{x}_{2,:} \\
    \vdots & \vdots & \ddots & \vdots \\
    \mathbf{y}_{1,:}^T\cdot\mathbf{x}_{m,:}  &  \mathbf{y}_{2,:}^T\cdot\mathbf{x}_{m,:} & \hdots & \mathbf{y}_{p,:}^T\cdot\mathbf{x}_{m,:} \\
     \end{bmatrix} \in \mathbb{R}^{qm \times pn}
\end{equation}
It is also important to keep in mind what each entry within $J_{\widehat{h}\circ\widehat{g}}$ mean:
\begin{center}
    $J_{\widehat{h}\circ\widehat{g}}=J_{\widehat{h}}\cdot J_{\widehat{g}} = 
    \begin{bmatrix}
    \frac{\partial c_{1,1}}{\partial b_{1,1}} & \frac{\partial c_{1,1}}{\partial b_{1,2}} &\hdots &  \frac{\partial c_{1,1}}{\partial b_{2,1}} &\hdots & \frac{\partial c_{1,1}}{\partial b_{m,p}} \\
    \frac{\partial c_{1,2}}{\partial b_{1,1}}  & \ddots &\hdots & \frac{\partial c_{1,2}}{\partial b_{2,1}}& \hdots& \vdots \\
    \vdots & \vdots & \ddots & \vdots & \hdots & \vdots\\
    \frac{\partial c_{2,1}}{\partial b_{1,1}} & \vdots & \vdots & \ddots & \hdots & \vdots\\
    \vdots & \vdots & \vdots & \vdots & \ddots & \vdots\\
    \frac{\partial c_{m,q}}{\partial b_{1,1}}  & \hdots & \hdots & \hdots & \hdots & \frac{\partial c_{m,q}}{\partial b_{m,p}}
\end{bmatrix} \cdot
    \begin{bmatrix}
    \frac{\partial b_{1,1}}{\partial a_{1,1}} & \frac{\partial b_{1,1}}{\partial a_{2,1}} &\hdots &  \frac{\partial b_{1,1}}{\partial a_{1,2}} &\hdots & \frac{\partial b_{1,1}}{\partial a_{n,p}} \\
    \frac{\partial b_{1, 2}}{\partial a_{1,1}}  & \ddots &\hdots & \frac{\partial b_{1,2}}{\partial a_{1,2}}& \hdots& \vdots \\
    \vdots & \vdots & \ddots & \vdots & \hdots & \vdots\\
    \frac{\partial b_{2,1}}{\partial a_{1,1}} & \vdots & \vdots & \ddots & \hdots & \vdots\\
    \vdots & \vdots & \vdots & \vdots & \ddots & \vdots\\
    \frac{\partial b_{m,p}}{\partial a_{1,1}}  & \hdots & \hdots & \hdots & \hdots & \frac{\partial b_{m,p}}{\partial a_{n,p}}
\end{bmatrix}
$
\end{center}
\begin{equation}
    = \begin{bmatrix}
    \frac{\partial c_{1,1}}{\partial a_{1,1}} & \frac{\partial c_{1,1}}{\partial a_{2,1}} &\hdots &  \frac{\partial c_{1,1}}{\partial a_{1,2}} &\hdots & \frac{\partial c_{1,1}}{\partial a_{n,p}} \\
    \frac{\partial c_{1,2}}{\partial a_{1,1}}  & \ddots &\hdots & \frac{\partial c_{1,2}}{\partial a_{1,2}}& \hdots& \vdots \\
    \vdots & \vdots & \ddots & \vdots & \hdots & \vdots\\
    \frac{\partial c_{2,1}}{\partial a_{1,1}} & \vdots & \vdots & \ddots & \hdots & \vdots\\
    \vdots & \vdots & \vdots & \vdots & \ddots & \vdots\\
    \frac{\partial c_{m,q}}{\partial a_{1,1}}  & \hdots & \hdots & \hdots & \hdots & \frac{\partial c_{m,q}}{\partial a_{n,p}}
\end{bmatrix}
\end{equation}

\subsection{Adjoint calculation}
Now we have successfully found the general solution of Jacobian matrix for $f(A)=XAY$ and $g(A)=XA$ (or more precisely, the unrolled version of them, namely $f(\mathbf{a})$ and $\widehat{g}(\mathbf{a})$, and it is time to apply them in the adjoint calculation for GCDE, as they are the special cases of the general solutions outlined in section 2.1-2.3. Back in section 1.2 I stated that we need to find the Jacobian matrix $\frac{\partial f(H(t), A, W)}{\partial H(t)}$ and $\frac{\partial f(H(t), A, W)}{\partial W}$; well that does not actually make sense in our convention since (1) $H(t)$ is not a vector and (2) the adjoint must be a vector, which means $\frac{\partial L}{\partial H(t)}$ should also be unrolled. Therefore, a GCDE version of adjoint dynamics for Equation (4) and (5) is:
\begin{equation}
    -a(t)^{T}\frac{\partial f(h(t), t, \theta)}{\partial h} \rightarrow 
    -\frac{\partial L}{\partial \mathbf{h}(t)}^{T}  \frac{\partial \widehat{f}(\mathbf{h}(t), A, W)}{\partial \mathbf{h}(t)} 
\end{equation}
\begin{equation}
    -a(t)^{T}\frac{\partial f(h(t), t, \theta)}{\partial \theta} \rightarrow
        -\frac{\partial L}{\partial \mathbf{h}(t)}^{T}  \frac{\partial \widehat{f}(\mathbf{h}(t), A, W)}{\partial \mathbf{w}}
\end{equation}
where $\mathbf{h}(t) = unroll(H(t))$, $\mathbf{w} = unroll(W)$, and $\widehat{f}(\mathbf{h}(t), A, W) = unroll(f(roll(\mathbf{h}(t)), A, W))$.
However, this does not stop us to find a vectorized equivalence for Equation (4) and (5) for GCDE implementation on hardware, as keeping the matrices unrolled greatly increases the dimensions and could not take the parallel in-memory computing advantages brought by memristor crossbars.

%The first adjoint, $\frac{\partial L}{\partial H(t)}^T\cdot\frac{\partial f(H(t), A, W)}{\partial H(t)}$

%$\frac{\partial f(H(t), A, W)}{\partial H(t)}$ and $\frac{\partial f(H(t), A, W)}{\partial W}$

\subsection{Vectorized Equation (22)}
We start the vectorization from the general case $f(A)=XAY$ we discussed in the earlier sections. Let $ \mathbb{R}^{m\times q}\ni C = f(A)$, and $\mathbb{R}^{mq}\ni\mathbf{c} = unroll(C)$, we unroll $C$ row by row so that its partial derivatives with respect to a scalar loss function $L$ has the form:
\begin{center}
    $\frac{\partial L}{\partial \mathbf{c}} = 
    \begin{bmatrix}
        \frac{\partial L}{\partial c_{1,1}} \\ \vdots \\\frac{\partial L}{\partial c_{1,q}}\\ \frac{\partial L}{\partial c_{2,1}} \\ \vdots \\ \frac{\partial L}{\partial c_{m,q}}
    \end{bmatrix} =
    \begin{bmatrix}
        \frac{\partial L}{\partial \mathbf{c}_{1,:}}^T \\ \frac{\partial L}{\partial \mathbf{c}_{2,:}}^T\\ \vdots \\ \frac{\partial L}{\partial \mathbf{c}_{m,:}}^T
    \end{bmatrix} \in \mathbb{R}^{mq}, \frac{\partial L}{\partial \mathbf{c}_{i,:}} \in \mathbb{R}^{1\times q}
    $ 
\end{center}
We want to find (a reminder that $\widehat{f}(\mathbf{a})$ is a unrolled version of $f(A)$):
\begin{center}
$
-\frac{\partial L}{\partial \mathbf{c}}^{T} \frac{\partial \widehat{f}(\mathbf{a})}{\partial \mathbf{a}}=
-\frac{\partial L}{\partial \mathbf{c}}^{T}  J_{\widehat{f}} = -\frac{\partial L}{\partial \mathbf{c}}^{T} J_{\widehat{h}\circ\widehat{g}}$
\end{center}
Expand and plug in Equation (20):
\begin{center}
    $ -\frac{\partial L}{\partial \mathbf{c}}^{T} J_{\widehat{h}\circ\widehat{g}} = 
    -\begin{bmatrix}
        \frac{\partial L}{\partial \mathbf{c}_{1,:}} & \frac{\partial L}{\partial \mathbf{c}_{2,:}} & \hdots & \frac{\partial L}{\partial \mathbf{c}_{m,:}}
    \end{bmatrix} \cdot
    \begin{bmatrix}
    \mathbf{y}_{1,:}^T\cdot\mathbf{x}_{1,:}  &  \mathbf{y}_{2,:}^T\cdot\mathbf{x}_{1,:} & \hdots & \mathbf{y}_{p,:}^T\cdot\mathbf{x}_{1,:} \\
    \mathbf{y}_{1,:}^T\cdot\mathbf{x}_{2,:}  &  \mathbf{y}_{2,:}^T\cdot\mathbf{x}_{2,:} & \hdots & \mathbf{y}_{p,:}^T\cdot\mathbf{x}_{2,:} \\
    \vdots & \vdots & \ddots & \vdots \\
    \mathbf{y}_{1,:}^T\cdot\mathbf{x}_{m,:}  &  \mathbf{y}_{2,:}^T\cdot\mathbf{x}_{m,:} & \hdots & \mathbf{y}_{p,:}^T\cdot\mathbf{x}_{m,:} \\
     \end{bmatrix}
    $
\end{center}
Since $\frac{\partial L}{\partial \mathbf{c}_{i,:}} \in \mathbb{R}^{1\times q}$ and $\mathbf{y}^T_{i,:}\cdot \mathbf{x}_{j,:} \in \mathbb{R}^{q\times n}$, the blocks defined above have matching dimensions. Hence we can take their dot product directly:
\begin{center}
    $ -\frac{\partial L}{\partial \mathbf{c}}^{T} J_{\widehat{h}\circ\widehat{g}} = -
    \begin{bmatrix}
        \frac{\partial L}{\partial \mathbf{c}_{1,:}} \cdot \mathbf{y}^T_{1,:}\cdot \mathbf{x}_{1,:}  + \hdots + \frac{\partial L}{\partial \mathbf{c}_{m,:}} \cdot \mathbf{y}^T_{1,:}\cdot \mathbf{x}_{m,:}  &
        \hdots & 
        \frac{\partial L}{\partial \mathbf{c}_{1,:}} \cdot \mathbf{y}^T_{p,:}\cdot \mathbf{x}_{1,:}   + \hdots + \frac{\partial L}{\partial \mathbf{c}_{m,:}} \cdot \mathbf{y}^T_{p,:}\cdot \mathbf{x}_{m,:} 
    \end{bmatrix}
    $
\end{center}
Note that this dot product is a long $1\times np$ vector, and we can simplify it by rolling it into a $p\times n$ matrix:

\begin{equation}
    -roll(\frac{\partial L}{\partial \mathbf{c}}^{T} J_{\widehat{h}\circ\widehat{g}}) = -
    \begin{bmatrix}
        \frac{\partial L}{\partial \mathbf{c}_{1,:}} \cdot \mathbf{y}^T_{1,:}\cdot \mathbf{x}_{1,:}   + \hdots + \frac{\partial L}{\partial \mathbf{c}_{m,:}} \cdot \mathbf{y}^T_{1,:}\cdot \mathbf{x}_{m,:} \\
        \frac{\partial L}{\partial \mathbf{c}_{1,:}} \cdot \mathbf{y}^T_{2,:}\cdot \mathbf{x}_{1,:}   + \hdots + \frac{\partial L}{\partial \mathbf{c}_{m,:}} \cdot \mathbf{y}^T_{2,:}\cdot \mathbf{x}_{m,:} \\
        \vdots\\
        \frac{\partial L}{\partial \mathbf{c}_{1,:}} \cdot \mathbf{y}^T_{p,:}\cdot \mathbf{x}_{1,:}   + \hdots + \frac{\partial L}{\partial \mathbf{c}_{m,:}} \cdot \mathbf{y}^T_{p,:}\cdot \mathbf{x}_{m,:} 
    \end{bmatrix}
\end{equation}
Equation (24) can be further simplified:
\begin{center}
    $-roll(\frac{\partial L}{\partial \mathbf{c}}^{T} J_{\widehat{h}\circ\widehat{g}}) = \begin{bmatrix}
        \frac{\partial L}{\partial \mathbf{c}_{1,:}} \cdot \mathbf{y}^T_{1,:} & \hdots & \frac{\partial L}{\partial \mathbf{c}_{m,:}} \cdot \mathbf{y}^T_{1,:} \\
        \frac{\partial L}{\partial \mathbf{c}_{1,:}} \cdot \mathbf{y}^T_{2,:}  &\hdots & \frac{\partial L}{\partial \mathbf{c}_{m, :}} \cdot \mathbf{y}^T_{2,:}\\
        \vdots & \ddots & \vdots\\
        \frac{\partial L}{\partial \mathbf{c}_{1,:}} \cdot \mathbf{y}^T_{p,:} &  \hdots & \frac{\partial L}{\partial \mathbf{c}_{m,:}} \cdot \mathbf{y}^T_{p,:} 
    \end{bmatrix} \cdot \begin{bmatrix}
        \mathbf{x}_{1,:} \\ \mathbf{x}_{2,:} \\ \vdots \\ \mathbf{x}_{m,:}
    \end{bmatrix} = \begin{bmatrix}
        \frac{\partial L}{\partial \mathbf{c}_{1,:}} \cdot \mathbf{y}^T_{1,:} & \hdots & \frac{\partial L}{\partial \mathbf{c}_{m,:}} \cdot \mathbf{y}^T_{1,:} \\
        \frac{\partial L}{\partial \mathbf{c}_{1,:}} \cdot \mathbf{y}^T_{2,:}  &\hdots & \frac{\partial L}{\partial \mathbf{c}_{m, :}} \cdot \mathbf{y}^T_{2,:}\\
        \vdots & \ddots & \vdots\\
        \frac{\partial L}{\partial \mathbf{c}_{1,:}} \cdot \mathbf{y}^T_{p,:} &  \hdots & \frac{\partial L}{\partial \mathbf{c}_{m,:}} \cdot \mathbf{y}^T_{p,:} 
    \end{bmatrix} \cdot X$
\end{center}
Let:
\begin{center}
    $E = \begin{bmatrix}
        \frac{\partial L}{\partial \mathbf{c}_{1,:}} \cdot \mathbf{y}^T_{1,:} & \hdots & \frac{\partial L}{\partial \mathbf{c}_{m,:}} \cdot \mathbf{y}^T_{1,:} \\
        \frac{\partial L}{\partial \mathbf{c}_{1,:}} \cdot \mathbf{y}^T_{2,:}  &\hdots & \frac{\partial L}{\partial \mathbf{c}_{m, :}} \cdot \mathbf{y}^T_{2,:}\\
        \vdots & \ddots & \vdots\\
        \frac{\partial L}{\partial \mathbf{c}_{1,:}} \cdot \mathbf{y}^T_{p,:} &  \hdots & \frac{\partial L}{\partial \mathbf{c}_{m,:}} \cdot \mathbf{y}^T_{p,:} 
    \end{bmatrix}$
\end{center}
We can simplify E further by looking at it row by row. The first row of E is:
\begin{equation}
    \mathbf{e}_{1,:} = \begin{bmatrix}
        \frac{\partial L}{\partial \mathbf{c}_{1,:}} \cdot \mathbf{y}^T_{1,:} & \hdots & \frac{\partial L}{\partial \mathbf{c}_{m,:}} \cdot \mathbf{y}^T_{1,:}
    \end{bmatrix}
\end{equation}
Notice that each entry within Equation (25) is a scalar because $\frac{\partial L}{\partial \mathbf{c}_{i,:}} \in \mathbb{R}^{1\times q}$ and $\mathbf{y}^T_{1,:} \in \mathbb{R}^{q\times 1}$. Therefore, we can manipulate Equation (25) such that: 
\begin{center}
    $\mathbf{e}_{1,:} = (\mathbf{y}^T_{1,:})^{T} \cdot \begin{bmatrix}
        \frac{\partial L}{\partial \mathbf{c}_{1,:}}^{T} & \frac{\partial L}{\partial \mathbf{c}_{2,:}}^{T} & \hdots &\frac{\partial L}{\partial \mathbf{c}_{m,:}}^{T} 
    \end{bmatrix} = \mathbf{y}_{1,:} \cdot \begin{bmatrix}
        \frac{\partial L}{\partial \mathbf{c}_{1,:}}^{T} & \frac{\partial L}{\partial \mathbf{c}_{2,:}}^{T} & \hdots &\frac{\partial L}{\partial \mathbf{c}_{m,:}}^{T} 
    \end{bmatrix}$
\end{center}
We define $\frac{\partial L}{\partial C}$ having matching entries to $C$:
\begin{center}
    $\frac{\partial L}{\partial C} = \begin{bmatrix}
        \frac{\partial L}{\partial c_{1,1}} & \frac{\partial L}{\partial c_{1,2}} & \hdots &\frac{\partial L}{\partial c_{1,q} } \\
        \frac{\partial L}{\partial c_{2,1}} & \ddots & \hdots &\vdots \\
        \vdots & \vdots & \ddots & \vdots\\
        \frac{\partial L}{\partial c_{m,1}} & \hdots & \hdots &\frac{\partial L}{\partial c_{m,q}}
    \end{bmatrix}$
\end{center}
Therefore:
\begin{center}
        $\mathbf{e}_{1,:} = \mathbf{y}_{1,:} \cdot \frac{\partial L}{\partial C}^T$
\end{center}
If we repeat for all rows of $E$, then:
\begin{center}
   $E = \begin{bmatrix}
        \mathbf{y}_{1,:} \cdot \frac{\partial L}{\partial C}^T \\
        \mathbf{y}_{2,:} \cdot \frac{\partial L}{\partial C}^T\\
        \vdots \\
        \mathbf{y}_{p,:} \cdot \frac{\partial L}{\partial C}^T
    \end{bmatrix}
    = \begin{bmatrix}
        \mathbf{y}_{1,:}  \\
        \mathbf{y}_{2,:} \\
        \vdots \\
        \mathbf{y}_{p,:} 
    \end{bmatrix} \cdot \frac{\partial L}{\partial C}^T = Y \cdot \frac{\partial L}{\partial C}^T$
\end{center}
As a result:
\begin{equation}
    -roll(\frac{\partial L}{\partial \mathbf{c}}^{T} J_{\widehat{h}\circ\widehat{g}}) = -E \cdot X = -Y \cdot \frac{\partial L}{\partial C}^T \cdot X
\end{equation}
We need to keep in mind what this matrix actually represents. So we go back to Equation (21):
\begin{center}
    $ -roll(\frac{\partial L}{\partial \mathbf{c}}^{T} J_{\widehat{h}\circ\widehat{g}}) = 
    -roll(\begin{bmatrix}
        \frac{\partial L}{\partial \mathbf{c}_{1,:}} & \frac{\partial L}{\partial \mathbf{c}_{2,:}} & \hdots & \frac{\partial L}{\partial \mathbf{c}_{m,:}}
    \end{bmatrix} \cdot= \begin{bmatrix}
    \frac{\partial c_{1,1}}{\partial a_{1,1}} & \frac{\partial c_{1,1}}{\partial a_{2,1}} &\hdots &  \frac{\partial c_{1,1}}{\partial a_{1,2}} &\hdots & \frac{\partial c_{1,1}}{\partial a_{n,p}} \\
    \frac{\partial c_{1,2}}{\partial a_{1,1}}  & \ddots &\hdots & \frac{\partial c_{1,2}}{\partial a_{1,2}}& \hdots& \vdots \\
    \vdots & \vdots & \ddots & \vdots & \hdots & \vdots\\
    \frac{\partial c_{2,1}}{\partial a_{1,1}} & \vdots & \vdots & \ddots & \hdots & \vdots\\
    \vdots & \vdots & \vdots & \vdots & \ddots & \vdots\\
    \frac{\partial c_{m,q}}{\partial a_{1,1}}  & \hdots & \hdots & \hdots & \hdots & \frac{\partial c_{m,q}}{\partial a_{n,p}}
\end{bmatrix})
    $
\end{center}

\begin{center}
$=-roll(\begin{bmatrix}
        \frac{\partial L}{\partial a_{1,1}} & \frac{\partial L}{\partial a_{2,1}} & \hdots & \frac{\partial L}{\partial a_{n,1}} & \frac{\partial L}{\partial a_{1,2}} & \hdots& \frac{\partial L}{\partial a_{n,p}}
    \end{bmatrix})$
\end{center}
\begin{equation}
=-\begin{bmatrix}
        \frac{\partial L}{\partial a_{1,1}} & \frac{\partial L}{\partial a_{2,1}} & \hdots & \frac{\partial L}{\partial a_{n,1}} \\
        \frac{\partial L}{\partial a_{1,2}} & \frac{\partial L}{\partial a_{2,2}} & \hdots & \frac{\partial L}{\partial a_{n,2}} \\
        \vdots & \vdots & \ddots & \vdots\\
        \frac{\partial L}{\partial a_{1,p}} & \frac{\partial L}{\partial a_{2,p}} & \hdots & \frac{\partial L}{\partial a_{n,p}} 
    \end{bmatrix}=
    -\begin{bmatrix}
        -\frac{\partial L}{\partial \mathbf{a}_{:,1}}^T- \\
        -\frac{\partial L}{\partial \mathbf{a}_{:,2}}^T- \\
        \vdots\\
        -\frac{\partial L}{\partial \mathbf{a}_{:,p}}^T- \\
    \end{bmatrix}
\end{equation}
Notice that the entries of this matrix matches the transpose of $A$. Hence, in order to match the entries of $A$, we will take Equation (27)'s transpose -- and here we found the general vectorized adjoint solution for $f(A)$:
\begin{center}
    $vectorized(-\frac{\partial L}{\partial \mathbf{c}}^{T} J_{\widehat{h}\circ\widehat{g}}) = -\begin{bmatrix}
        -\frac{\partial L}{\partial \mathbf{a}_{:,1}}^T- \\
        -\frac{\partial L}{\partial \mathbf{a}_{:,2}}^T- \\
        \vdots\\
        -\frac{\partial L}{\partial \mathbf{a}_{:,p}}^T- \\
    \end{bmatrix}^T = -(Y \cdot \frac{\partial L}{\partial C}^T \cdot X)^T $
\end{center}
\begin{equation}
    vectorized(-\frac{\partial L}{\partial \mathbf{c}}^{T} J_{\widehat{h}\circ\widehat{g}}) = -X^T \cdot \frac{\partial L}{\partial C} \cdot Y^T
\end{equation}
The adjoint dynamics shown in Equation (22) for GCDE is just a special of Equation (28). The derivative of the $ReLU()$ activation function is an element-wise binary step function:
\begin{equation}
    step(x)=
    \begin{cases} 
      1 & x>0 \\
      0 & x\leq 0 
   \end{cases}
\end{equation}
Hence, let $\odot$ denote the Hadamard product between matrices, by chain rule and Equation (28), the vectorized Equation (22) is:
\begin{center}
    $vectorized(-\frac{\partial L}{\partial \mathbf{h}(t)}^{T}  \frac{\partial \widehat{f}(\mathbf{h}(t), A, W)}{\partial \mathbf{h}(t)}) = 
    -A^T \cdot (\frac{\partial L}{\partial H(t)} \odot step(H(t)))\cdot W^T$
\end{center}
where:
\begin{center}
    $\frac{\partial L}{\partial H(t)} = roll(\frac{\partial L}{\partial \mathbf{h}(t)})=\begin{bmatrix}
        \frac{\partial L}{\partial h_{1,1}} & \frac{\partial L}{\partial h_{1,2}} & \hdots &\frac{\partial L}{\partial h_{1,q} } \\
        \frac{\partial L}{\partial h_{2,1}} & \ddots & \hdots &\vdots \\
        \vdots & \vdots & \ddots & \vdots\\
        \frac{\partial L}{\partial h_{m,1}} & \hdots & \hdots &\frac{\partial L}{\partial h_{m,q}}
    \end{bmatrix}$
\end{center}
Since by definition of GCDE, the graph topology matrix A is always symmetric, the final vectorized Equation (22) is:
\begin{equation}
 vectorized(-\frac{\partial L}{\partial \mathbf{h}(t)}^{T}  \frac{\partial \widehat{f}(\mathbf{h}(t), A, W)}{\partial \mathbf{h}(t)}) = 
    -A \cdot (\frac{\partial L}{\partial H(t)} \odot step(H(t)))\cdot W^T
\end{equation}
\subsection{Vectorized Equation (23)}
We will again vectorize Equation (23) from a general case, $g(A)=XA$. This is because we can treat $AH(t)$ in Equation (6) as a single matrix that linearly transforms $W$. Let $ \mathbb{R}^{m\times p}\ni B = g(A)$, and $\mathbb{R}^{mp}\ni\mathbf{b} = unroll(B)$, we unroll $B$ row by row so that its partial derivatives with respect to a scalar loss function $L$ has the form:
\begin{center}
    $\frac{\partial L}{\partial \mathbf{b}} = 
    \begin{bmatrix}
        \frac{\partial L}{\partial b_{1,1}} \\ \vdots \\\frac{\partial L}{\partial b_{1,p}}\\ \frac{\partial L}{\partial b_{2,1}} \\ \vdots \\ \frac{\partial L}{\partial b_{m,p}}
    \end{bmatrix} =
    \begin{bmatrix}
        \frac{\partial L}{\partial \mathbf{b}_{1,:}}^T \\ \frac{\partial L}{\partial \mathbf{b}_{2,:}}^T\\ \vdots \\ \frac{\partial L}{\partial \mathbf{b}_{m,:}}^T
    \end{bmatrix} \in \mathbb{R}^{mp}, \frac{\partial L}{\partial \mathbf{b}_{i,:}} \in \mathbb{R}^{1\times p}
    $ 
\end{center}
In addition, we define
\begin{center}
    $\frac{\partial L}{\partial B} = \begin{bmatrix}
        \frac{\partial L}{\partial b_{1,1}} & \frac{\partial L}{\partial b_{1,2}} & \hdots &\frac{\partial L}{\partial b_{1,q} } \\
        \frac{\partial L}{\partial b_{2,1}} & \ddots & \hdots &\vdots \\
        \vdots & \vdots & \ddots & \vdots\\
        \frac{\partial L}{\partial b_{m,1}} & \hdots & \hdots &\frac{\partial L}{\partial b_{m,q}}
    \end{bmatrix}$
\end{center}
Since we are considering the general case $g(A)=XA$ and we are given an arbitury adjoint $\mathbf{b}$, we can change Equation (17) to:
\begin{center}
$-\frac{\partial L}{\partial \mathbf{h}(t)}^{T}  \frac{\partial \widehat{f}(\mathbf{h}(t), A, W)}{\partial \mathbf{w}} \rightarrow 
-\frac{\partial L}{\partial \mathbf{b}}^{T} \frac{\partial \widehat{g}(\mathbf{a})}{\partial \mathbf{a}} = -\frac{\partial L}{\partial \mathbf{b}}^{T}  J_{\widehat{g}}$
\end{center}
Expand and plug in Equation (17):
\begin{center}
    $ -\frac{\partial L}{\partial \mathbf{b}}^{T} J_{\widehat{g}} = 
    -\begin{bmatrix}
        \frac{\partial L}{\partial \mathbf{b}_{1,:}} & \frac{\partial L}{\partial \mathbf{b}_{2,:}} & \hdots & \frac{\partial L}{\partial \mathbf{b}_{m,:}}
    \end{bmatrix} \cdot
    \begin{bmatrix}
     \mathbf{x}_{1,:}  & &   &  \\
      & \mathbf{x}_{1,:} & &  \\
     &  & \ddots &\\
     &  &  & \mathbf{x}_{1,:}\\
     & \hdots& \hdots &\\
     \mathbf{x}_{m,:}  & &   &  \\
      & \mathbf{x}_{m,:} & &  \\
     &  & \ddots &\\
     &  &  & \mathbf{x}_{m,:}
\end{bmatrix}$
\end{center}
We again break it into blocks:
\begin{center}
    $ -\frac{\partial L}{\partial \mathbf{b}}^{T} J_{\widehat{g}} = 
    -\begin{bmatrix} \begin{bmatrix}
        \frac{\partial L}{\partial \mathbf{b}_{1,:}} & \frac{\partial L}{\partial \mathbf{b}_{2,:}} & \hdots & \frac{\partial L}{\partial \mathbf{b}_{m,:}}
    \end{bmatrix} \cdot
    \begin{bmatrix}
     \mathbf{x}_{1,:}  \\
       \vdots\\
    \textbf{0} \\
     \mathbf{x}_{2,:} \\
      \vdots\\
     \mathbf{x}_{m,:} \\
    \vdots\\
    \textbf{0} \\
\end{bmatrix} \hdots 
    \begin{bmatrix}
        \frac{\partial L}{\partial \mathbf{b}_{1,:}} & \frac{\partial L}{\partial \mathbf{b}_{2,:}} & \hdots & \frac{\partial L}{\partial \mathbf{b}_{m,:}}
    \end{bmatrix} \cdot
    \begin{bmatrix}
      \textbf{0} \\
       \vdots\\
    \mathbf{x}_{1,:} \\
      \textbf{0} \\
      \vdots\\
      \textbf{0}\\
    \vdots\\
    \mathbf{x}_{m,:} \\
\end{bmatrix}
\end{bmatrix}$
\end{center}
If we expand the first block:
\begin{center}
    $\begin{bmatrix}
        \frac{\partial L}{\partial \mathbf{b}_{1,:}} & \frac{\partial L}{\partial \mathbf{b}_{2,:}} & \hdots & \frac{\partial L}{\partial \mathbf{b}_{m,:}}
    \end{bmatrix} \cdot
    \begin{bmatrix}
     \mathbf{x}_{1,:}  \\
       \vdots\\
    \textbf{0} \\
     \mathbf{x}_{2,:} \\
      \vdots\\
     \mathbf{x}_{m,:} \\
    \vdots\\
    \textbf{0} \\
\end{bmatrix}  = \begin{bmatrix}
 \frac{\partial L}{\partial b_{1,1}} \cdot x_{1,1} + \frac{\partial L}{\partial b_{2,1}} \cdot x_{2,1} + \hdots + \frac{\partial L}{\partial b_{m,1}} \cdot x_{m,1} \\
 \frac{\partial L}{\partial b_{1,1}} \cdot x_{1,2} + \frac{\partial L}{\partial b_{2,1}} \cdot x_{2,2} + \hdots + \frac{\partial L}{\partial b_{m,1}} \cdot x_{m,2} \\
 \vdots\\
 \frac{\partial L}{\partial b_{1,1}} \cdot x_{1,p} + \frac{\partial L}{\partial b_{2,1}} \cdot x_{2,p} + \hdots + \frac{\partial L}{\partial b_{m,1}} \cdot x_{m,p} 
\end{bmatrix}^T$
\end{center}
\begin{center}
    $= (\begin{bmatrix}
  x_{1,1} &  x_{2,1} & \hdots &  x_{m,1} \\
  x_{1,2} &  x_{2,2} & \hdots &  x_{m,2} \\
 \vdots &\vdots& \ddots &\vdots\\
  x_{1,p} &  x_{2,p} & \hdots &  x_{m,p} 
\end{bmatrix} \cdot \begin{bmatrix}
 \frac{\partial L}{\partial b_{1,1}} \\ \frac{\partial L}{\partial b_{2,1}} \\ \vdots \\ \frac{\partial L}{\partial b_{m,1}}
\end{bmatrix})^T = (X^T \cdot \frac{\partial L}{\partial \mathbf{b}_{:,1}})^T$
\end{center}
If we repeat for other blocks, we would get:
\begin{center}
    $ -\frac{\partial L}{\partial \mathbf{b}}^{T} J_{\widehat{g}} = \begin{bmatrix}
    X^T \cdot \frac{\partial L}{\partial \mathbf{b}_{:,1}} \\ 
    X^T \cdot \frac{\partial L}{\partial \mathbf{b}_{:,2}} \\
    \vdots\\
    X^T \cdot \frac{\partial L}{\partial \mathbf{b}_{:,p}}
    \end{bmatrix}^T$ or $ (-\frac{\partial L}{\partial \mathbf{b}}^{T} J_{\widehat{g}})^T = \begin{bmatrix}
    X^T \cdot \frac{\partial L}{\partial \mathbf{b}_{:,1}} \\ 
    X^T \cdot \frac{\partial L}{\partial \mathbf{b}_{:,2}} \\
    \vdots\\
    X^T \cdot \frac{\partial L}{\partial \mathbf{b}_{:,p}}
    \end{bmatrix}$
\end{center}
We will make it vectorized by rolling it into a $n\times p$ matrix:
\begin{center}
    $roll((-\frac{\partial L}{\partial \mathbf{b}}^{T} J_{\widehat{g}})^T) = -\begin{bmatrix}
    X^T \cdot \frac{\partial L}{\partial \mathbf{b}_{:,1}} & 
    X^T \cdot \frac{\partial L}{\partial \mathbf{b}_{:,2}} &
    \hdots&
    X^T \cdot \frac{\partial L}{\partial \mathbf{b}_{:,p}}
    \end{bmatrix}$
\end{center}
Further vectorization could be done:
\begin{center}
    $-\begin{bmatrix}
    X^T \cdot \frac{\partial L}{\partial \mathbf{b}_{:,1}} & 
    X^T \cdot \frac{\partial L}{\partial \mathbf{b}_{:,2}} &
    \hdots&
    X^T \cdot \frac{\partial L}{\partial \mathbf{b}_{:,p}}
    \end{bmatrix} = -X^T \cdot \begin{bmatrix}
     \frac{\partial L}{\partial \mathbf{b}_{:,1}} & 
     \frac{\partial L}{\partial \mathbf{b}_{:,2}} &
    \hdots&
     \frac{\partial L}{\partial \mathbf{b}_{:,p}}
    \end{bmatrix}$
\end{center}
\begin{equation}
    = -X^T \cdot \frac{\partial L}{\partial B}
\end{equation}
Of course, we need to keep in mind what $-X^T \cdot \frac{\partial L}{\partial B}$ actually represents:
\begin{center}
    $roll((-\frac{\partial L}{\partial \mathbf{b}}^{T} J_{\widehat{g}})^T) =  -(\begin{bmatrix}
        \frac{\partial L}{\partial b_{1,1}} & \hdots &\frac{\partial L}{\partial b_{1,p}}& \frac{\partial L}{\partial b_{2,1}} & \hdots & \frac{\partial L}{\partial b_{m,p}}
    \end{bmatrix} \cdot \begin{bmatrix}
    \frac{\partial b_{1,1}}{\partial a_{1,1}} & \frac{\partial b_{1,1}}{\partial a_{2,1}} &\hdots &  \frac{\partial b_{1,1}}{\partial a_{1,2}} &\hdots & \frac{\partial b_{1,1}}{\partial a_{n,p}} \\
    \frac{\partial b_{1, 2}}{\partial a_{1,1}}  & \ddots &\hdots & \frac{\partial b_{1,2}}{\partial a_{1,2}}& \hdots& \vdots \\
    \vdots & \vdots & \ddots & \vdots & \hdots & \vdots\\
    \frac{\partial b_{2,1}}{\partial a_{1,1}} & \vdots & \vdots & \ddots & \hdots & \vdots\\
    \vdots & \vdots & \vdots & \vdots & \ddots & \vdots\\
    \frac{\partial b_{m,p}}{\partial a_{1,1}}  & \hdots & \hdots & \hdots & \hdots & \frac{\partial b_{m,p}}{\partial a_{n,p}}
\end{bmatrix})^T$
\end{center}
\begin{center}
$=-roll(\begin{bmatrix}
        \frac{\partial L}{\partial a_{1,1}} & \frac{\partial L}{\partial a_{2,1}} & \hdots & \frac{\partial L}{\partial a_{n,1}} & \frac{\partial L}{\partial a_{1,2}} & \hdots& \frac{\partial L}{\partial a_{n,p}}
    \end{bmatrix}^T)$
\end{center}
\begin{equation}
=-\begin{bmatrix}
        \frac{\partial L}{\partial a_{1,1}} & \frac{\partial L}{\partial a_{1,2}} & \hdots & \frac{\partial L}{\partial a_{1,p}} \\
        \frac{\partial L}{\partial a_{2,1}} & \frac{\partial L}{\partial a_{2,2}} & \hdots & \frac{\partial L}{\partial a_{2,p}} \\
        \vdots & \vdots & \ddots & \vdots\\
        \frac{\partial L}{\partial a_{n,1}} & \frac{\partial L}{\partial a_{n,2}} & \hdots & \frac{\partial L}{\partial a_{n,p}} 
    \end{bmatrix}
\end{equation}
\begin{center}
    $ = -X^T \cdot \frac{\partial L}{\partial B}$
\end{center}
Since the entries of $-X^T\cdot \frac{\partial L}{\partial B}$ matches $A$, Equation (31) is a good vectorized solution. Therefore:
\begin{center}
    $vectorized(-\frac{\partial L}{\partial \mathbf{b}}^{T}  J_{\widehat{g}}) = -X^T\cdot \frac{\partial L}{\partial B}$
\end{center}
Now if we plug Equation (31) into our GCDE special case, and if we define:
\begin{center}
    $\frac{\partial L}{\partial W} = roll(\frac{\partial L}{\partial \mathbf{w}})=\begin{bmatrix}
        \frac{\partial L}{\partial w_{1,1}} & \frac{\partial L}{\partial w_{1,2}} & \hdots &\frac{\partial L}{\partial w_{1,q} } \\
        \frac{\partial L}{\partial w_{2,1}} & \ddots & \hdots &\vdots \\
        \vdots & \vdots & \ddots & \vdots\\
        \frac{\partial L}{\partial w_{m,1}} & \hdots & \hdots &\frac{\partial L}{\partial w_{m,q}}
    \end{bmatrix}$
\end{center}
Then the vectorized Equation (23) is:
\begin{equation}
 vectorized(-\frac{\partial L}{\partial \mathbf{w}}^{T}  \frac{\partial \widehat{f}(\mathbf{h}(t), A, W)}{\partial \mathbf{w}}) = 
    -(AH(t))^T \cdot (\frac{\partial L}{\partial W} \odot step(H(t)))
\end{equation}
\subsection{Summary}
In conclusion, we have found the vectorized adjoint dynamics for GCDE. Using the vectorized adjoint dynamics, we do not need to unroll the hidden states and parameters, and we could take the advantages of the in-memory matrices programmed on memristor crossbar.
\section*{Acknowledgement}
I thank Louis Primeau for his thorough tutorial on multivariable calculus and matrix unrolling for Jacobian calculation. I thank Nafiseh Ghoroghchian for her guide on matrix differentiation and proofread of the work.
\bibliographystyle{alpha}
\bibliography{sample}
[1]  R. T. Q. Chen, Y. Rubanova, J. Bettencourt, \& D. Duvenaud, ‘Neural Ordinary Differential Equations’. arXiv, 2018. https://arxiv.org/pdf/1806.07366.pdf \\
$[2]$ T. N. Kipf \& M. Welling, ‘Semi-Supervised Classification with Graph Convolutional Networks’. arXiv, 2016. https://arxiv.org/pdf/1911.07532.pdf \\
$[3]$
M. Poli, S. Massaroli, J. Park, A. Yamashita, H. Asama, \& J. Park, ‘Graph Neural Ordinary Differential Equations’. arXiv, 2019. https://arxiv.org/abs/1911.07532\\
$[4]$ Louis Primeau, Neural ODE for Memristor Crossbar.
\end{document}